\pdfoutput=1
\documentclass[10pt,twocolumn,letterpaper]{article}

 \usepackage{cvpr}              

\usepackage[dvipsnames]{xcolor}
\usepackage{tabu}
\usepackage{multirow}
\usepackage{amsmath,bm}
\usepackage{adjustbox}
\usepackage{color,xcolor}
\usepackage{colortbl}
\usepackage{bbding}
\usepackage[accsupp]{axessibility}

\definecolor{cvprblue}{rgb}{0.21,0.49,0.74}

\usepackage[pagebackref,breaklinks,colorlinks,citecolor=cvprblue]{hyperref}

\DeclareMathOperator{\mlp}{MLP}

\DeclareMathOperator{\concat}{Concat}

\DeclareMathOperator{\conv}{Conv}
\DeclareMathOperator{\dist}{Dist}
\DeclareMathOperator{\gaussian}{Gaussian}

\def\paperID{3680} 
\def\confName{CVPR}
\def\confYear{2024}

\title{DPMesh: Exploiting Diffusion Prior for Occluded Human Mesh Recovery}

\def\spaces{~~~~~~}
\author{Yixuan Zhu\textsuperscript{1}\thanks{Equal contribution. ~~\textsuperscript{\dag}Corresponding authors.}\spaces{}Ao Li\textsuperscript{2}\footnotemark[1]\spaces{}Yansong Tang$^{2\dagger}$\spaces{}Wenliang Zhao\textsuperscript{1}\spaces{}Jie Zhou\textsuperscript{1}\spaces{}Jiwen Lu\textsuperscript{1}\\\\
\textsuperscript{1}Department of Automation, Tsinghua University \\
\textsuperscript{2}Tsinghua Shenzhen International Graduate School, Tsinghua University
}

\begin{document}
\maketitle
\begin{abstract}
The recovery of occluded human meshes presents challenges for current methods due to the difficulty in extracting effective image features under severe occlusion. 
In this paper, we introduce DPMesh, an innovative framework for occluded human mesh recovery that capitalizes on the profound diffusion prior about object structure and spatial relationships embedded in a pre-trained text-to-image diffusion model. 
Unlike previous methods reliant on conventional backbones for vanilla feature extraction, DPMesh seamlessly integrates the pre-trained denoising U-Net with potent knowledge as its image backbone and performs a single-step inference to provide occlusion-aware information. To enhance the perception capability for occluded poses, DPMesh incorporates well-designed guidance via condition injection, which produces effective controls from 2D observations for the denoising U-Net. Furthermore, we explore a dedicated noisy key-point reasoning approach to mitigate disturbances arising from occlusion and crowded scenarios. This strategy fully unleashes the perceptual capability of the diffusion prior, thereby enhancing accuracy. Extensive experiments affirm the efficacy of our framework, as we outperform state-of-the-art methods on both occlusion-specific and standard datasets. The persuasive results underscore its ability to achieve precise and robust 3D human mesh recovery, particularly in challenging scenarios involving occlusion and crowded scenes. 
Code is available at \url{https://github.com/EternalEvan/DPMesh}.
\end{abstract}  
\section{Introduction}
\label{sec:intro}
\begin{figure}[t]
  \centering

    \includegraphics[width=0.99\linewidth]{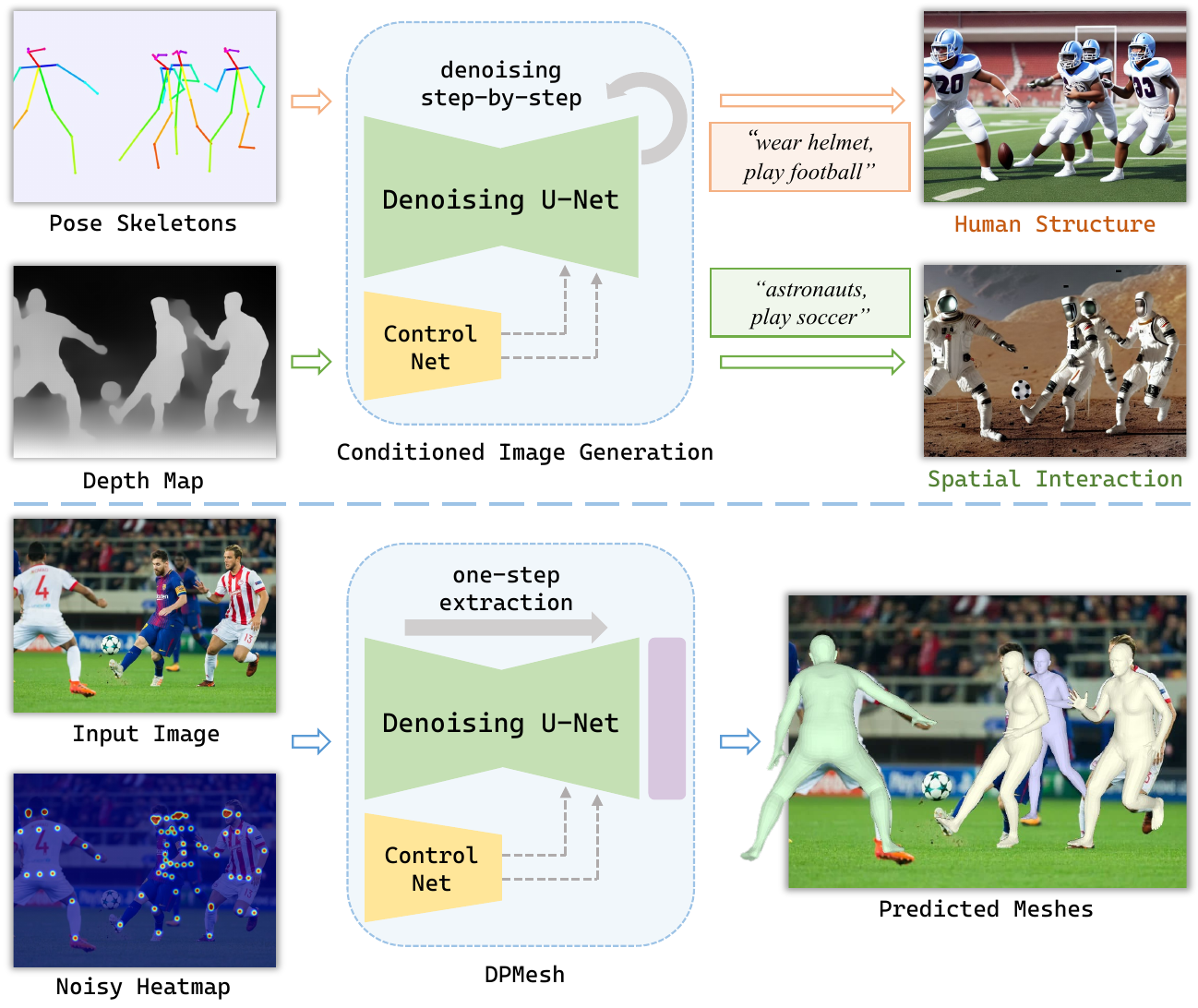}
 \caption{\textbf{Main idea of the proposed \textit{DPMesh} framework.} We design an innovative framework to fully exploit rich prior knowledge about human structure and spatial interaction of the pre-trained diffusion model for challenging occluded human mesh recovery task. By simply adapting the denoising U-Net as a single-step backbone with spatial conditions, we achieve accurate human mesh recovery even under severe occlusions. 
 }
  \label{fig:teaser}
  \vspace{-10pt}
\end{figure}
\begin{figure*}
  \centering

    \includegraphics[width=0.99\linewidth]{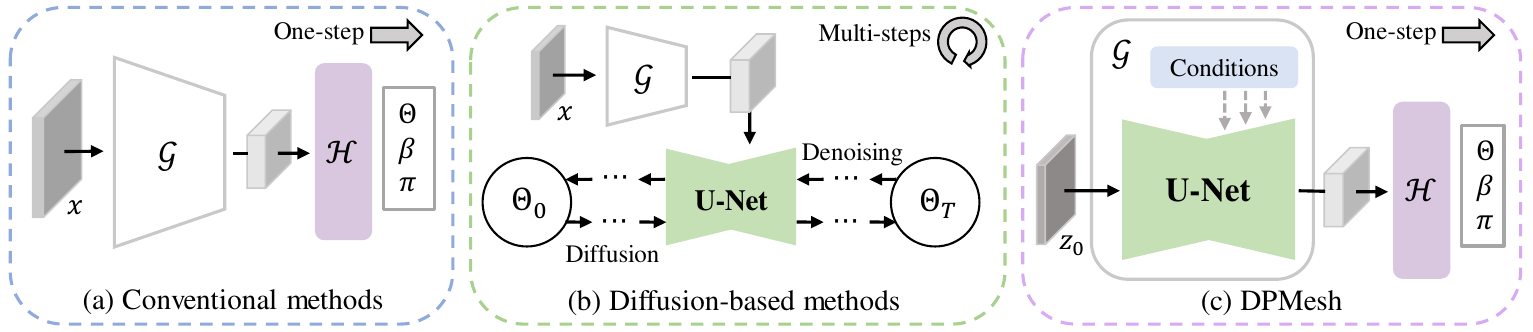}
    \caption{\textbf{Comparison of current methods and the proposed \textit{DPMesh}.} (a) Conventional methods~\cite{hmr,kolotouros2019learning} apply a feature extractor $\mathcal{G}$ and a regressor $\mathcal{H}$ to obtain SMPL parameters. (b) Diffusion-based methods~\cite{feng2023diffpose,cho2023generative,hmdiff2023distribution} propose an iterative framework that harnesses multiple denoising steps to progressively refine the pose parameters from random noise. (c) Distinct from previous diffusion-based techniques, our DPMesh employs the pre-trained denoising U-Net as the backbone $\mathcal{G}$, executing a one-step inference to furnish informative features for the regressor. This novel framework transfers the potent perception knowledge in generative models onto conventional frameworks.}
  \label{fig:fig2}
  \vspace{-10pt}
\end{figure*}
The goal of human mesh recovery involves estimating the 3D human pose and shape from either monocular or multi-view images and videos. Over the past decade, this field has evolved into a burgeoning and captivating research problem, gaining prominence for its extensive applications in film-making, game development, and sports. In recent years, a plethora of approaches~\cite{hmr,romp,kolotouros2019learning,pavlakos2018learning,choutas2020monocular,choi2020pose2mesh,kocabas2020vibe,kocabas2021pare,lin2021end,sun2022putting,zhang2021pymaf} grounded in deep learning have emerged, paving the way to address this inherently ill-posed problem by effectively regressing body parameters from image features.

Nonetheless, extracting more effective information from monocular images in complex scenarios (\textit{e.g.}, occlusions and crowded environments), remains a pivotal challenge. Existing methods address the intricacies of mesh recovery in complex scenarios by incorporating 2D prior knowledge as hints, drawing the models' attention to visible body parts and reinforcing their 2D alignment proficiency. Following this line, mainstream methods~\cite{crowdnet,li2023jotr} intuitively apply off-the-shelf key-point detectors to achieve coarse human joints, others such as~\cite{kocabas2021pare} and~\cite{3doh} introduce partial segmentation masks and UV maps as pixel-level knowledge. Despite these efforts, persistent shortcomings become apparent, particularly when confronted with severe occlusion, since they excessively depend on 2D alignment and disregard the vivid information embedded in natural images. Consequently, disturbances to the 2D detector due to noise and occlusion significantly impact accuracy, yielding unsatisfactory outcomes. Recently, Diffusion Models (DMs)~\cite{ho2020denoising,rombach2022high} have introduced a step-by-step generation framework, showcasing remarkable image synthesis capacity and producing visually appealing results. Inspired by DMs, recent works like~\cite{cho2023generative,feng2023diffpose,hmdiff2023distribution} utilize the generative approach proposed by diffusion models for pose estimation, achieving high-accuracy results. However, diffusion-based methods suffer from repeated iterations and neglect the learned knowledge for image processing within text-to-image diffusion models, causing the potential of diffusion not fully exploited. 
More recent studies~\cite{zhao2023unleashing,zhang2023adding,wang2024prolificdreamer} investigate the pre-trained diffusion model for 3D-related tasks, \textit{e.g.}, image synthesis from the depth map, text-to-3D generation and depth estimation. It has been verified that the pre-trained diffusion model can provide structure-aware knowledge for 3D generation and perception tasks. Although diffusion models possess rich knowledge of 3D structure and spatial interaction from generative training, the challenge persists in effectively leveraging these capabilities for complex regression tasks like occluded human mesh recovery.

To overcome the aforementioned challenges, we present DPMesh, a simple yet effective framework for occluded human mesh recovery. DPMesh employs a pre-trained text-to-image diffusion model as the backbone, fully leveraging its potent knowledge of the 3D structure and spatial relationships learned from generative training, hence yielding a robust estimator for occluded poses, as illustrated in Figure~\ref{fig:teaser}. Our primary goal is to harness both the high-level and low-level visual concepts within a pre-trained diffusion model
for the demanding occluded pose estimation task. Instead of following the time-consuming step-by-step denoising process, we replace conventional image backbones with the pre-trained denoising U-Net and perform an efficient single-step inference style with designed conditions as guidance, as depicted in Figure~\ref{fig:fig2}. Considering the pre-training of the diffusion model on text-to-image generation tasks, we confront two challenges: (1) preserving the learned knowledge within the pre-trained diffusion model and adapting it to the occluded human mesh recovery task, and (2) designing appropriate conditions and controls to enhance the model's perception ability. To address these issues, we introduce an efficient framework to tailor the diffusion model for mesh recovery, leveraging an effective condition injection. 
To align with the original diffusion model and facilitate the interaction between image features and 2D prior information, we refine the spatial information from an off-the-shelf detector and inject the diffusion model with these conditions. This yields detailed knowledge of the 2D position and the key-points uncertainty. 
The processed 2D information serves as guidance for the diffusion model, ultimately producing rich visual content, encompassing both human structure and spatial interaction for the subsequent regressor. 
Furthermore, we present a noisy key-point reasoning approach to improve the robustness of our model, rendering it more stable for occlusion and crowds.



We conduct extensive experiments on various occlusion benchmarks 3DPW-OC~\cite{3dpw,3doh}, 3DPW-PC~\cite{3dpw,romp}, 3DOH~\cite{3doh}, 3DPW-Crowd~\cite{3dpw,crowdnet} and CMU-Panoptic~\cite{cmu}, as well as the standard benchmark 3DPW test split~\cite{3dpw}. Remarkably, without any finetuning on the 3DPW training set, our DPMesh achieves an exciting performance, surpassing previous state-of-the-art methods and demonstrating significantly improved accuracy. Specifically, we achieve MPJPE values of 70.9, 82.2, 79.9, and 73.6 on 3DPW-OC, 3DPW-PC, 3DPW-Crowd, and 3DPW test split, respectively, underscoring the proficiency of our framework.
Furthermore, we carry out comprehensive ablation studies to highlight the effectiveness of the diffusion-based backbone, the condition construction and the designed noisy key-point reasoning. 

\section{Related Work}
\label{sec:relate work}
\noindent \textbf{Human Mesh Recovery.} During the past decade, parameterized human model~\cite{anguelov2005scape,loper2015smpl,SMPL-X:2019} has been widely used to express 3D human pose and shape. Many proceeding works explore approaches to estimate accurate model parameters from monocular images~\cite{kolotouros2019learning,hmr,pavlakos2018learning,guler2018densepose,guler2019holopose,zanfir2020weakly,choutas2020monocular,choi2020pose2mesh,kocabas2020vibe,kocabas2021pare,lin2021end,sun2022putting,zhang2021pymaf}. 
They usually regress parameters from extracted image features. Some~\cite{zhang2021pymaf,ma20233d} leverage 2D and 3D visual observations to enhance the 2D alignment of image features. Nevertheless, they always fall short when confronted with complex scenarios, \textit{e.g.} occlusion and crowded environments since the conventional backbones and estimators provide vague information about the occluded region for the regressor. To handle this challenge, a series of methods~\cite{li2023jotr,kocabas2021pare,choi2022learning,3doh,khirodkar2022occluded,sun2021monocular} propose effective approaches involving segmentation masks, center maps and 3D representations to improve the 2D and 3D alignments. However, they have limitations since they pay too much attention to enhancing the usage of the 2D and 3D observations and ignore the quality of image features, which are fundamental and significant. Most recently, methods based on diffusion models~\cite{cho2023generative,feng2023diffpose,hmdiff2023distribution} introduce an iterative framework to estimate human poses in the repeated denoising process. Though they achieve satisfying accuracy, they suffer from extensive time-consuming and do not fully exploit the rich knowledge in diffusion models.

\noindent\textbf{Diffusion Models.}
Diffusion denoising probabilistic models, commonly referred to as diffusion models~\cite{ho2020denoising,rombach2022high}, have emerged as a prominent family of generative models, showcasing remarkable synthesis quality and controllability. The core concept of the diffusion model involves training a denoising autoencoder to estimate the inverse of a Markovian diffusion process~\cite{sohl2015deep}. Through generative training on large-scale datasets 
with image-text pairs (\textit{e.g.}, LAION-5B~\cite{schuhmann2022laion}), diffusion models acquire a powerful capability to generate high-quality images with diverse content and reasonable structures. This proficiency is harnessed during diffusion sampling, which can be perceived as a progressive denoising procedure that necessitates repeated inference of the denoising autoencoder. Recently,~\cite{zhang2023adding} propose a controllable architecture, named ControlNet, to add spatial controls, \textit{e.g.,} depth maps and human poses, to pre-trained diffusion models, broadening their applications to controlled image generation. Although originally tailored to 2D text-to-image tasks, pre-trained diffusion models also possess rich knowledge about object structure and spatial interaction. They can adapt to various 3D-related tasks like image synthesis from depth map, text-to-3D generation and depth estimation, as explored in~\cite{zhao2023unleashing,zhang2023adding,wang2024prolificdreamer}. However, fully exploiting the structure-aware generative prior in the diffusion model for complex mesh recovery, especially in occluded and crowded scenarios, still poses a significant challenge due to the need for proficient visual perception capability.

\begin{figure*}
  \centering
    \includegraphics[width=0.99\linewidth]{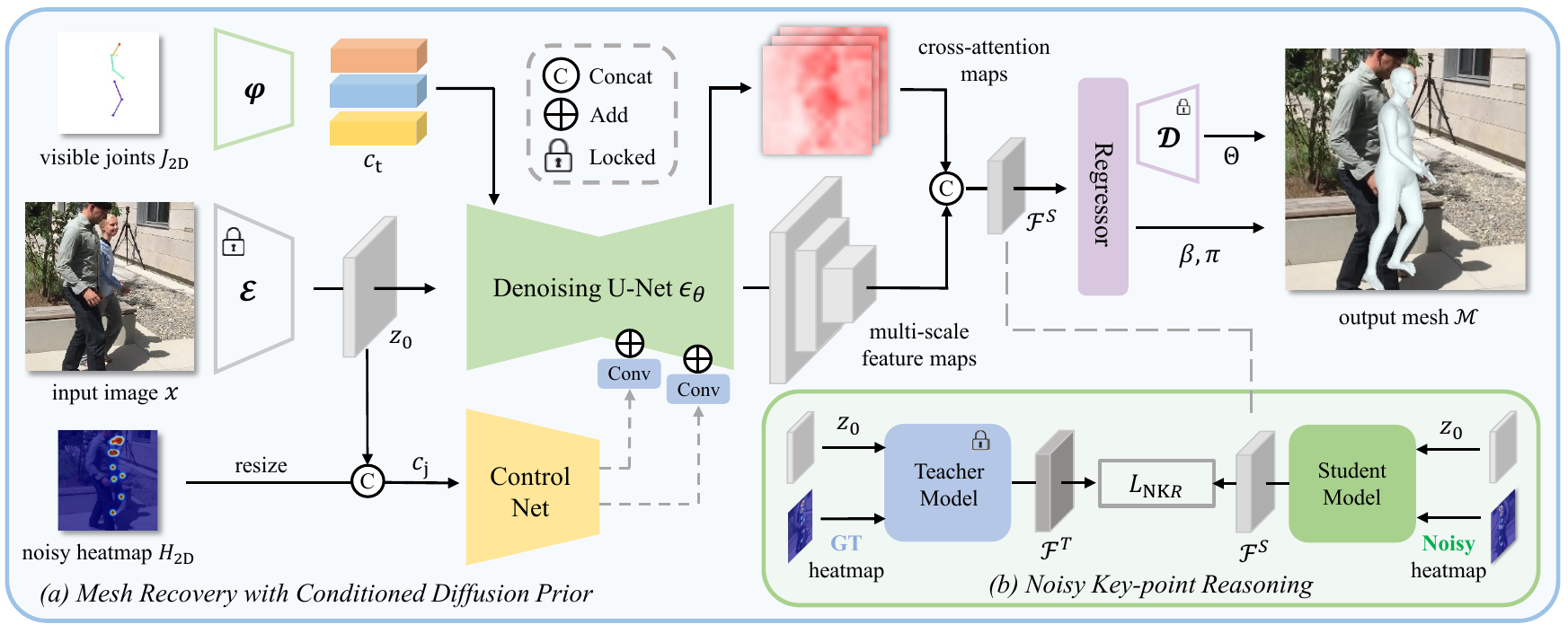}
  \vspace{-10pt}
    \caption{\textbf{The overall framework of \textit{DPMesh}.} Given the input image ${\bm x}$, pre-detected 2D key-points $J_{\rm 2D}$, and generated heatmap $H_{\rm 2D}$, our framework begins with the extraction of image features $\mathcal{F}^S$ through a single denoising step using the pre-trained diffusion model. This process is guided by the designed spatial conditions. Then, we input $\mathcal{F}^S$ to the regressor to predict SMPL parameters $\Theta$, $\beta$ and $\pi$, ultimately generating the final mesh. To further enhance the estimation robustness against noisy 2D observations, we leverage a noisy key-point reasoning approach. This involves pre-training a teacher model with ground truth heatmaps and then training a student model using noisy heatmaps by aligning the feature maps computed from the locked teacher and the student.}
  \label{fig:pipeline}
  \vspace{-10pt}
\end{figure*}

\section{Methods}
\label{sec:methods}
In this section, we present DPMesh, an effective framework for occluded human mesh recovery with the pre-trained diffusion prior and proper conditional control.
We will start by reviewing the background of diffusion models with conditional control and the human body model. Then we will provide a detailed walkthrough of the entire pipeline to introduce our designs in DPMesh. This includes how we leverage the generative diffusion prior for the human recovery task and inject valuable conditions to guide the denoising U-Net. Moreover, we will present a noisy key-point reasoning approach to enhance the robustness of our model. The overall framework of our DPMesh is illustrated in Figure~\ref{fig:pipeline}.

\subsection{Preliminaries} 
\label{subsec: preliminaries}
\noindent
\textbf{Conditional Control for Diffusion Models.} Diffusion models achieve high controllability thanks to the effective cross-attention layers in the denoising U-Net ${\bm \epsilon}_\theta$~\cite{ronneberger2015u} which bridges a way for the interactions between image features and various conditions. Recently, ControlNet~\cite{zhang2023adding} successfully enhances the fine-grained spatial control on latent diffusion model (LDM)~\cite{rombach2022high} by leveraging a trainable copy of the encoding layers in the denoising U-Net as a strong backbone for learning diverse conditional controls. During the training of the ControlNet framework, images are first projected to latent representations ${\bm z}_0$ by a trained VQGAN consisting of the encoder $\mathcal E$ and the decoder $\mathcal D$. Denoting ${\bm z}_t$ as the noisy image at $t$-th timestep, it is produced by:
\begin{equation}
  {\bm z}_t = \sqrt{\bar{\alpha}_t}{\bm z}_0+\sqrt{1-\bar{\alpha}_t}{\bm \epsilon},
  \label{eq:preliminaries latent diffusion 1}
\end{equation}
where $\bar{\alpha}_t=\prod_{s=1}^{t}\alpha_s$ and ${\bm \epsilon} \sim \mathcal{N}(0,{\bm I})$. Given the noisy image and conditions, the training objective of the ControlNet framework can be derived as: 
\begin{equation}
  L_{\rm CLDM}=\mathbb{E}_{{\bm z}_0,t,{\bm c}_{\rm t},{\bm c}_{\rm f},{\bm \epsilon}} \bigg[\Vert {\bm \epsilon}-{\bm \epsilon}_{\theta}({\bm z}_t,t,{\bm c}_{\rm t},{\bm c}_{\rm f}) \Vert_2^2 \bigg],
  \label{eq:preliminaries latent diffusion 2}
\end{equation}
where ${\bm z}_t$ is computed from Equation (\ref{eq:preliminaries latent diffusion 1}), ${\bm c}_{\rm t}$ denotes the text condition embedding extracted from frozen CLIP~\cite{radford2021learning} text encoder and ${\bm c}_{\rm f}$ is a task-specific condition, such as human skeleton poses or canny maps. To prevent harmful noise that influences the hidden states of neural network layers at the start of training, ControlNet applies zero convolution layers for the trainable copy branch. The condition branch consumes ${\bm c}_{\rm f}$ as input and injects the outcomes to the output blocks of diffusion model ${\bm \epsilon}_\theta$. In order to keep generation capability and reduce computational costs, it freezes the parameters in ${\bm \epsilon}_\theta$. By utilizing the fine-grained conditions, ControlNet successfully achieves controllable human image generation with various conditions like 2D skeletons.

\noindent
\textbf{Human Body Model.} We use SMPL~\cite{loper2015smpl} model to parameterize human body mesh. SMPL represents 3D human body with three vectors, denoted by pose $\Theta \in \mathbb{R}^{72}$, shape $\beta \in \mathbb{R}^{10}$ and camera parameters $\pi \in \mathbb{R}^{3}$. The body mesh is generated by a differentiable function $\mathcal{M}(\Theta,\beta) \in \mathbb{R}^{6890}$. Then we can obtain the 3D joint coordinates by $\mathcal{J}_{\rm 3D}=\mathcal{W}\mathcal{M} \in \mathbb{R}^{N \times 3}$, where $\mathcal{W}$ is a pre-trained linear regressor and $N$ represents the number of joints. With the predicted camera parameters $\pi$, we can obtain reprojected 2D joints $\mathcal{J}_{\rm 2D}=\Pi(\mathcal{J}_{\rm 3D},\pi) \in \mathbb{R}^{N \times 2}$ by perspective projection. 

\subsection{Diffusion Prior for Occluded Mesh Recovery}

\noindent
\textbf{Overview.} Our primary goal is to fully exploit the pre-trained diffusion model's potential for occluded human mesh recovery, leveraging its learned knowledge of the object structure and spatial interactions. In contrast to previous methods involving repeated diffusion sampling, our basic idea is to simply employ the pre-trained diffusion model as the image backbone, performing a single inference to extract features from image ${\bm x}$. To provide effective guidance, we adopt the condition injection to play an essential role in processing pre-detected 2D observations into the conditions. Then we utilize the noisy key-point reasoning approach to improve the occlusion awareness of our model to further enhance the robustness of the proposed framework. In conclusion, we propose DPMesh, which takes the image and corresponding noisy key-point observations as inputs and estimates the SMPL parameters $ \Theta$, $\beta$ and $\pi$, collectively denoted as the output ${\bm y}$. This process can be formulated as:
\begin{equation}
  p_{\phi}({\bm y}|{\bm x}) = p_{\phi_3}({\bm y}|\mathcal{F})p_{\phi_2}(\mathcal{F}|{\bm x},\mathcal{C})
   p_{\phi_1}(\mathcal{C}|{\bm x}),
  \label{eq:overview}
\end{equation}
where $\mathcal{F}$ denotes the extracted feature maps and $\mathcal{C}$ represents the conditions. We will describe the role of each term in Equation (\ref{eq:overview}) along with their detailed designs.

\noindent
\textbf{Condition Injection with 2D HeatMap.} $p_{\phi_1}(\mathcal{C}|{\bm x})$ aims to construct effective conditions, which is significant since it provides spatial guidance for the denoising U-Net backbone ${\bm \epsilon}_{\theta}$. Therefore we design the condition injection to introduce high-level and spatial information that guides the backbone to focus on the region of interest. For each cropped image, we utilize an off-the-shelf 2D key-point detector~\cite{cao2017realtime} to obtain 2D joints $J_{\rm 2D}$ along with their corresponding confidence and generate the heatmaps $H_{\rm 2D} \in \mathbb{R}^{N \times H_{0} \times W_{0}}$ as conditioning input using 2-dimensional Gaussian kernels, where $N$ represents the number of detected key-points. After that, we concatenate the heatmap $H_{2D}$ with the input image ${\bm z}_0$ to obtain ${\bm c}_{\rm j} \in \mathbb{R}^{(N+C_{z}) \times H_0 \times W_0}$. It is noteworthy that, in the original ControlNet architecture, the conditioning image is fused with ${\bm z}_0$ by element-wise adding after passing through convolution layers. However, we observe that this addition significantly damages the final condition quality. The preparation for ${\bm c}_{\rm j}$ can be expressed as:
\begin{equation}
  {\bm c}_{\rm j} = \concat({\bm z}_0,H_{\rm 2D}), H_{\rm 2D} = \gaussian(J_{\rm 2D}).
  \label{eq:controlnet}
\end{equation}
Then we employ the ControlNet architecture to process fine-grained conditions from ${\bm c}_{\rm j}$ and inject them to the image features in the denoising U-Net ${\bm \epsilon}_\theta$. The output of the $i$-th layer $F_i$ in the decoding layers of ${\bm \epsilon}_\theta$ is derived as:
\begin{equation}
  F_i = F(F_{i-1};\theta_i)+\conv(F({\bm c}_{\rm j};\theta_{\rm cond});\theta_{\rm Conv}),
\end{equation}
where $F_{i-1}$ is the output of the previous block and $F(\cdot;\theta)$ denotes a trained neural network. $\theta_i,\theta_{\rm cond}$ represent the parameters within the denoising U-Net and ControlNet, respectively. $\theta_{\rm Conv}$ is the parameters of zero convolution layers with both weights and bias initialized to zeros. Note that in the original ControlNet architecture, $\theta_{\rm cond}$ is a trainable copy of the encoding blocks in the denoising U-Net.

Besides the ControlNet that provides controls for the denoising U-Net ${\bm \epsilon}_\theta$, we also consider using the cross-attention prompt to pinpoint more accurate spatial information. In original diffusion models, the prompt ${\bm c}_{\rm t}$ conditions typically consist of text embeddings from frozen CLIP. However, in our framework, we replace text with visible 2D joint coordinates $J_{\rm 2D} \in \mathbb{R}^{N \times 2}$. We then apply a two-layer MLP to elevate the dimension of 2D joint coordinates to match the text token dimension $D_{\rm t}$, which is set to 768 in the pre-trained diffusion model. Thus we obtain the auxiliary spatial condition ${\bm c}_{\rm t} \in \mathbb{R}^{N \times D_t}$. We take ${\bm c}_{\rm t}$ as prompt guidance and send it to all cross-attention blocks in the denoising U-Net. The construction of ${\bm c}_{\rm t}$ can be formulated as:
\begin{equation}
  {\bm c}_{\rm t} = \mlp(J_{\rm 2D}).
\end{equation}
To sum up, we construct the condition set $\mathcal{C}$ consisting of ${\bm c}_{\rm j}$ and ${\bm c}_{\rm t}$ and inject them into ${\bm \epsilon}_\theta$ through different passways.

\noindent
\textbf{Feature Extraction with Diffusion Prior.} $p_{\phi_2}(\mathcal{F}|{\bm x},\mathcal{C})$ is dedicated to the feature extraction from the input image. In contrast to previous methods that employ convolution-based and Transformer-based backbones, we leverage the pre-trained diffusion model as our backbone, exploiting the visual perception capability within the denoising U-Net learned from the generative training. As verified in~\cite{zhao2023unleashing}, there exist enough visual priors about the object structure and spatial interactions in a pre-trained denoising U-Net ${\bm \epsilon}_\theta$. By unlocking the potential of ${\bm \epsilon}_\theta$, we can tune the generative capability in the pre-trained diffusion model to address the human mesh recovery task. Motivated by this perspective, we design a straightforward feature extractor implemented by the pre-trained denoising U-Net, which receives effective guidance $\mathcal{C}$ from the condition injection. To prepare the input images, we convert the cropped image ${\bm x} \in \mathbb{R}^{H \times W \times C}$ from pixel space to the latent space with the frozen encoder $\mathcal E$ to obtain the latent representation ${\bm z}_0 \in \mathbb{R}^{H_{0} \times W_{0} \times C_{z}}$. Then we feed ${\bm z}_0$ into the pre-trained U-Net ${\bm \epsilon}_{\theta}$ and extract the multi-scale feature maps $F_i$, where $i \in \{1,4,7\}$, from the decoding layers as the implicit image features. Furthermore, we empirically observe that the cross-attention maps $T_i \in \mathbb{R}^{|{\bm c}_{\rm t}| \times H_i \times W_i}$ can provide significant occlusion-aware information indicating the invisible parts and explicit object structure knowledge about the human pose and shape. Therefore we concatenate the cross-attention maps with the feature maps $F_i$ and obtain the hierarchical feature maps $\mathcal{F}\leftarrow \{[F_i, T_i]\}$, which incorporate both implicit and explicit diffusion priors for subsequent regression.

\noindent
\textbf{SMPL Mesh Regressor.} $p_{\phi_3}({\bm y}|\mathcal{F})$ represents the regressor responsible for predicting the parameters of the body model from feature maps $\mathcal{F}$. To capture the body information in $\mathcal{F}$, we employ a cascade Transformer decoder for the regressor. In order to provide sufficient human pose priors and maintain the symmetry of the VQGAN framework, we train a VQVAE on a large motion dataset~\cite{AMASS:2019} with massive SMPL pose parameters to learn discrete representations for human poses. During the final regression, we predict the entry indices of the learned codebook and feed the corresponding pose embedding to the decoder $\mathcal{D}$ of the VQVAE to attain the pose parameters $\Theta$. As for shape and camera parameters (\textit{i.e.}\ $\beta$ and $\pi$), which are highly dependent on image features, we directly regress them using linear layers. 
\begin{table*}[t]
	\centering
 \caption{ \textbf{Quantitative comparisons on occlusion benchmarks.} The units for mean joint and vertex errors are in millimeters. Our DPMesh demonstrates outstanding estimation accuracy across diverse occlusion conditions, underscoring the efficacy of our framework.}
     \vspace{-10pt}
 \adjustbox{width=\linewidth}
	{\begin{tabular}{lcccccccccccc} 
	\toprule 
	\multirow{2}{*}{Method}& \multicolumn{3}{c}{3DPW-OC} &\multicolumn{3}{c}{3DPW-PC} &\multicolumn{3}{c}{3DOH}
 &\multicolumn{3}{c}{3DPW-Crowd}\\
 \cmidrule(lr){2-4}\cmidrule(lr){5-7}\cmidrule(lr){8-10}\cmidrule(lr){11-13}
	&MPJPE$\downarrow$ &PA-MPJPE$\downarrow$ &MPVE$\downarrow$   &MPJPE$\downarrow$ &PA-MPJPE$\downarrow$ &MPVE$\downarrow$    &MPJPE$\downarrow$ &PA-MPJPE$\downarrow$ &MPVE$\downarrow$ 
 &MPJPE$\downarrow$ &PA-MPJPE$\downarrow$ &MPVE$\downarrow$\\
    \midrule
        {SPIN~\cite{kolotouros2019learning}}&95.5 & 60.7&121.4 &122.1&77.5&159.8 &110.5&71.6&124.2
        &121.2 &69.9 &144.1\\
	{PyMAF~\cite{zhang2021pymaf}}&89.6 & 59.1&113.7& 117.5&74.5&154.6 &101.6&67.7&116.6& 115.7 &66.4 &147.5\\
        {ROMP~\cite{romp}}&91.0& 62.0&-&98.7&69.0&-&-&-&-&104.8 &63.9 &127.8\\{OCHMR~\cite{khirodkar2022occluded}}&112.2& 75.2&145.9&- &-&-&-&-&-&-&-&-\\
        {PARE~\cite{kocabas2021pare}}&83.5& 57.0&101.5 &95.8&64.5&122.4 &109.0&63.8&117.4&94.9 &57.5 &117.6\\
        {3DCrowdNet~\cite{crowdnet}}&83.5& 57.1&101.5 &90.9&64.4&114.8 &102.8&61.6&111.8&85.8 &55.8 &108.5\\
       {JOTR~\cite{li2023jotr}}&75.7& 52.2&92.6 &86.5&58.3&109.7 &98.7&59.3&\textbf{104.8}&82.4 &52.0 &103.4\\
    
    \midrule
    \rowcolor{gray!30}DPMesh (Ours)& \textbf{70.9} & \textbf{48.0} & \textbf{88.0}  &\textbf{82.2}& \textbf{56.6} & \textbf{105.4}&\textbf{97.1}&\textbf{59.0}&106.4
    &\textbf{79.9}&\textbf{51.1}&\textbf{101.5}\\
	
	\bottomrule 
	\end{tabular}}
    
    \label{tab: main experiments}
    \vspace{-10pt}
\end{table*}

\subsection{Noisy Key-point Reasoning}
As we introduce an off-the-shelf 2D key-point detector for providing 2D observation hints, there naturally arises a problem: \textit{How robust is our framework in the presence of noisy key-points?} The noisy key-points, often arising from severe occlusion, can adversely impact the model's performance during evaluation. This consideration motivates us to reinforce our backbone with extra supervision to mitigate the model's reliance on noisy key-points. To achieve this, we leverage a self-supervised distillation approach, called \textit{Noisy Key-point Reason (NKR)}, that focuses on 2D detection errors, including missing key-points, jitters and mismatch. The core concept involves training a teacher model adept at accurately encoding feature maps with precise ground truth key-points. Then we utilize the teacher's feature maps $\mathcal{F}^{T}$
to guide and supervise the student's feature maps $\mathcal{F}^{S}$. During the distillation process, we minimize both the SimCLR loss~\cite{chen2020simple} and MSE loss:
\begin{equation}
    L_{\rm NKR} = L_{\rm SimCLR} + L_{\rm MSE},
  \label{eq:simclr}
\end{equation}
where the SimCLR loss is computed by:
\begin{equation}
     L_{\rm SimCLR} = \log \frac{\exp(\dist(F_{i}^{S}, F_{i}^{T}))}{\sum_{i \neq j} \exp(\dist(F_{i}^{S}, F_{j}^{T}))}.
\end{equation}
The SimCLR loss extends the distance between the two models' features for different inputs while minimizing the distance for the same input. The MSE loss also provides additional supervision to align features between teacher and student. This noisy key-point reasoning approach enhances the robustness of our framework against 2D detection errors, ensuring its stability under challenging occlusion.

\subsection{Implementation}
We use Stable Diffusion V1-5~\cite{rombach2022high} with ControlNet~\cite{zhang2023adding}, pre-trained for human-like image generation from 2D skeletons, as our image backbone. Following~\cite{crowdnet,li2023jotr}, we take the cropped image in 256 $\times$ 256 resolution as input and encode it into the latent code ${\bm z}_0 \in \mathbb{R}^{4 \times 32 \times 32}$. We extract feature maps with the size 8, 16 and 32 and the cross-attention maps in the same resolution from the denoising U-Net. To maintain the learned knowledge in the pre-trained diffusion model, we use LoRA~\cite{hu2022lora} to unfreeze the linear layers in cross-attention blocks, setting the rank of LoRA modules to 64. We find that even fine-tuning a small number of parameters via LoRA yields satisfying results. More details are shown in the appendix file.

Finally, we obtain mesh vertices $\mathcal{M}(\Theta,\beta) \in \mathbb{R}^{6890}$ and 3D joints $\mathcal{J}_{\rm 3D}=\mathcal{W}\mathcal{M} \in \mathbb{R}^{N \times 3}$ with functions mentioned in \ref{subsec: preliminaries}. We reproject the body joints to the image by $\mathcal{J}_{\rm 2D}=\Pi(\mathcal{J}_{\rm 3D},\pi) \in \mathbb{R}^{N \times 2}$. In the reprojection process, we approximately estimate the focal length with the length of the image diagonal following~\cite{li2022cliff}. For the training objectives, we utilize wide-used losses on SMPL parameters, 2D joints and 3D joints when 3D joint annotations are available to optimize our framework. In conclusion, the entire loss function can be formulated as: 
\begin{equation}
\begin{split}
  L =&\ \lambda_{\rm 2D}L_{\rm 2D} + \lambda_{\rm 3D}L_{\rm 3D} + \\
  &\ \lambda_{\rm SMPL}L_{\rm SMPL} + \lambda_{\rm NKR}L_{\rm NKR}.
\end{split}
  \label{eq:whole loss}    
\end{equation}
The first three terms are computed by:
\begin{equation}
    L_{\rm 2D} = \Vert \mathcal{J}_{\rm 2D} - \hat{\mathcal{J}}_{\rm 2D} \Vert , \ \ L_{\rm 3D} = \Vert \mathcal{J}_{\rm 3D} - \hat{\mathcal{J}}_{\rm 3D} \Vert ,
  \label{eq:split loss 1}    
\end{equation}
\begin{equation}
    L_{\rm SMPL} = \Vert \Theta - \hat{\Theta} \Vert + \Vert \beta - \hat{\beta} \Vert ,
  \label{eq:split loss 2}    
\end{equation}
where $\hat{\mathcal{J}}_{\rm 2D}$, $\hat{\mathcal{J}}_{\rm 3D}$, $\hat{\Theta}$ and $\hat{\beta}$ represent the ground truth annotations of 2D joints, 3D joints, SMPL body pose parameters and shape parameters, respectively.

\begin{figure*}[t]
  \centering

    \includegraphics[width=0.97\linewidth]{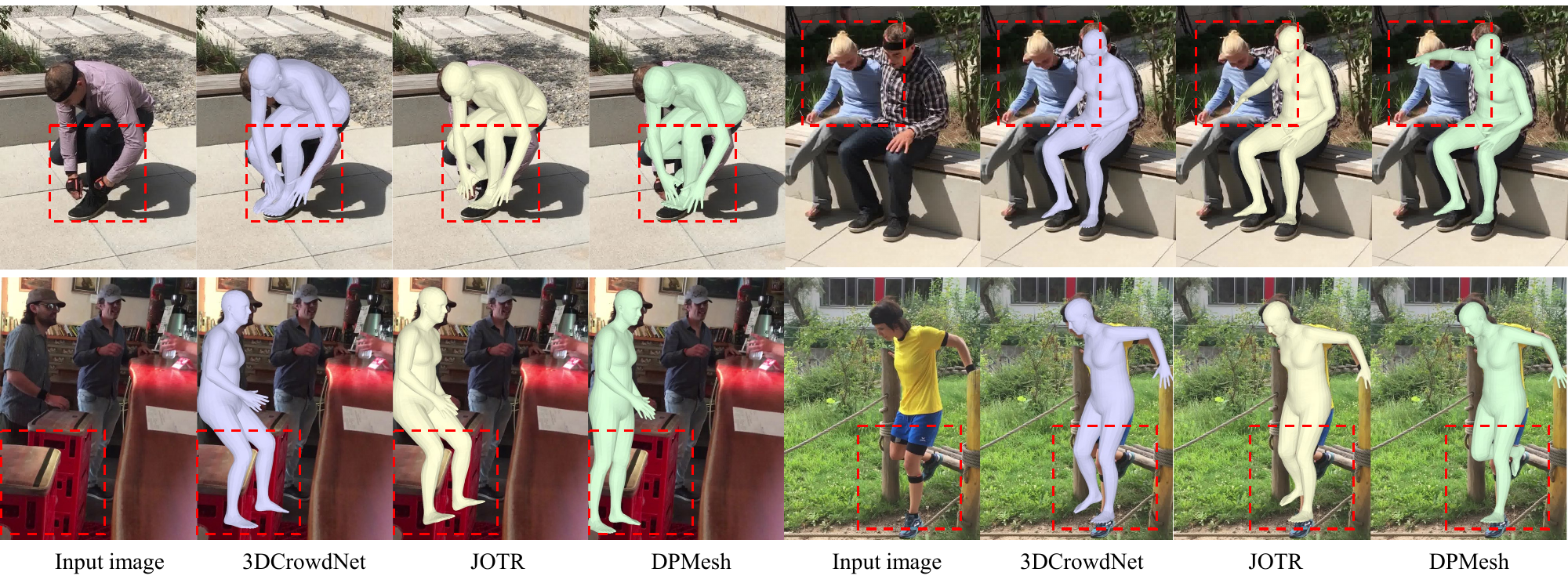}
  \vspace{-10pt}
    \caption{\textbf{Qualitative comparisons on 3DPW dataset~\cite{3dpw}.} Our DPMesh recovers accurate human meshes under challenging occlusions and demonstrates an adept understanding of 3D body structures and spatial relationships. Notably, our method also excels in generating plausible details for the obscured body parts, \textit{e.g.}, hands and legs, proving the robustness of DPMesh in handling complex scenarios.  }
  \label{fig:performance1}
  \vspace{-10pt}
\end{figure*}

    

\section{Experiments}
To verify the effectiveness of our proposed DPMesh, we conduct comprehensive experiments and ablation studies on the standard benchmark and various occlusion datasets. We will introduce the experimental settings including the implementation for training and evaluation. Subsequently, we will present our main results and offer a comprehensive analysis through detailed ablations.
\label{sec: experiments}

\subsection{Experiment Setup}
\noindent
\textbf{Training Details.}
In alignment with previous works~\cite{crowdnet,li2023jotr}, we train our model on a hybrid dataset with 2D or 3D annoations, including Human3.6M~\cite{ionescu2013human3}, MuCo-3DHP~\cite{mehta2018single}, MSCOCO~\cite{lin2014microsoft}, and CrowdPose~\cite{li2019crowdpose}. We exclusively utilize the training sets of these datasets, adhering to standard split protocols. For the 2D dataset, we utilize their pseudo ground-truth SMPL parameters~\cite{moon2022neuralannot}. During training, we add realistic errors on the ground truth (GT) 2D pose following~\cite{moon2019posefix,choi2020pose2mesh} to simulate erroneous 2D pose, rather than generating detected key-point results. 
We use AdamW optimizer with a batch size of 16 and a weight decay of 1e-6.We set the initial learning rate to 1e-4 and cut it to 1e-5 in the last 5 epochs. 
We adopt the AdamW optimizer, paired with a batch size of 16 and a weight decay of 1e-6. The whole training process takes 30 epochs. The initial learning rate is set at 1e-4, which is reduced to 1e-5 for the final 5 epochs. The VQVAE is trained on AMASS~\cite{AMASS:2019} for 2,000 epochs and the teacher model is trained with ground truth key-point labels for 10 epochs.

\noindent
\textbf{Evaluation Details.} We evaluate our model on 3DPW~\cite{3dpw} test split, 3DOH~\cite{3doh} test split, 3DPW-PC~\cite{3dpw,romp}, 3DPW-OC~\cite{3dpw,3doh}, 3DPW-Crowd~\cite{3dpw,crowdnet} and CMU-Panoptic dataset~\cite{cmu}. 3DPW-PC is the \textit{person-person} occlusion subset of 3DPW and 3DPW-OC is the \textit{person-object} occlusion subset of 3DPW. 3DOH is another \textit{person-object} occlusion dataset. The metrics we use are mean per joint position error (MPJPE) in mm, Procrustes-aligned mean per joint position error (PA-MPJPE) in mm for evaluating the accuracy of 3D joints and mean per vertex error (MPVE) in mm for evaluating 3D mesh error. For the CMU-Panoptic dataset, we only evaluate MPJPE following previous work~\cite{hmr,fieraru2021remips,zanfir2018monocular,zanfir2018deep}.

\subsection{Comparisons on Occlusion Benchmark}
\label{subsec: occlusion}

\noindent
\textbf{3DPW-OC}~\cite{3dpw,3doh} is a person-object occlusion subset of 3DPW and contains 20243 persons. Table~\ref{tab: main experiments} shows that our DPMesh outperforms all competitors with 70.9 mm MPJPE and 48.0 mm PA-MPJPE, demonstrating its promising capability in effectively handling complex in-the-wild scenes.

\begin{figure}[t]
  \centering

    \includegraphics[width=0.97\linewidth]{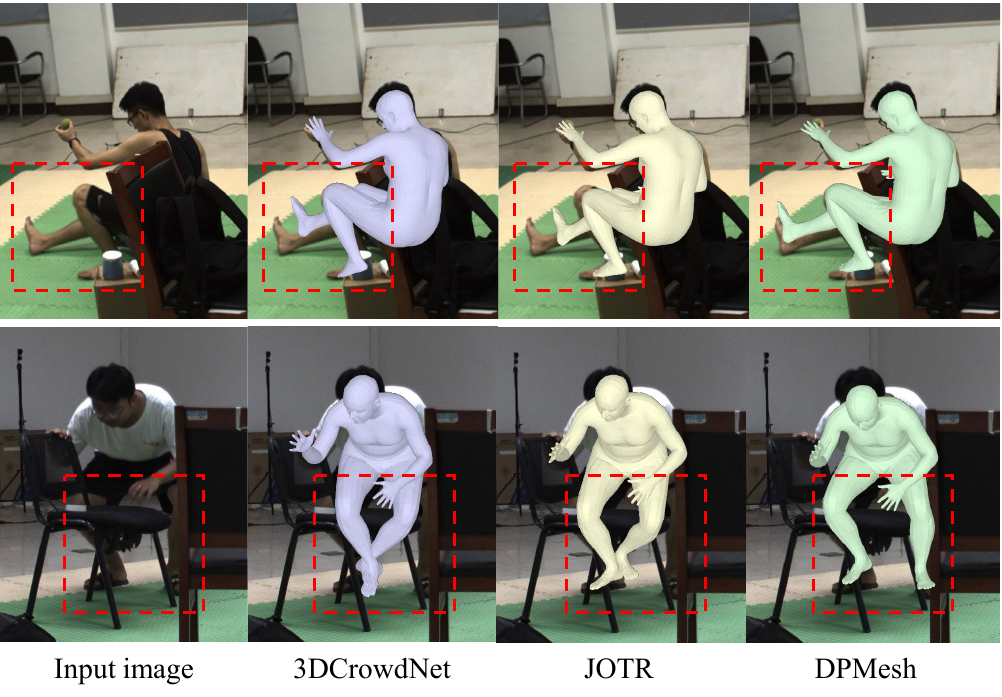}
      \vspace{-10pt}
 \caption{\textbf{Qualitative results on 3DOH dataset~\cite{3dpw}.} Our DPMesh obtains satisfying estimation for complex poses.}
  \label{fig:oh3d}
  \vspace{-10pt}
\end{figure}

\noindent
\textbf{3DOH}~\cite{3doh} is a person-object occlusion-specific dataset that encompasses 1290 images in the test split. All methods we report are not fine-tuned on the training split for fair comparison. We achieve the best results on 97.1 MPJPE and 59.0 PA-MPJPE, as shown in Table~\ref{tab: main experiments}. We further exhibit the qualitative findings in Figure~\ref{fig:oh3d}. Our DPMesh proficiently manages heavy occlusion situations.

\begin{table}[t]
\centering
\small
\caption{ \textbf{Quantitative comparisons on 3DPW~\cite{3dpw} test split.} DPMesh seamlessly generalizes to previously unseen distributions, yielding robust results on real-world RGB videos.}
  \vspace{-10pt}
      \begin{tabular}{lcccc}
      \toprule
        Method    & MPJPE$\downarrow$  & PA-MPJPE$\downarrow$ & MPVE$\downarrow$\\
        \midrule
        {HMR}~\cite{hmr}  &130.0 &76.7 & - \\
        {GraphCMR}~\cite{gcmr} &- &70.2 & -\\
        {SPIN}~\cite{kolotouros2019learning} & 96.9 & 56.2 & 116.4 \\
        {PyMaf}~\cite{zhang2021pymaf} & 92.8 & 58.9 & 110.1 \\
        {OCHMR}~\cite{khirodkar2022occluded} & 89.7 & 58.3 & 107.1 \\
        {ROMP}~\cite{romp} & 89.3 & 53.5 & 105.6 \\
        {PARE}~\cite{kocabas2021pare} & 82.9 & 52.3 & 99.7 \\
        {3DCrowdNet}~\cite{crowdnet} & 81.7 & 51.2 & 98.3 \\
        {JOTR}~\cite{li2023jotr} & 76.4 & 48.7 & 92.6 \\
        \midrule
        \rowcolor{gray!30}DPMesh (Ours) & \textbf{73.6} & \textbf{47.4} & \textbf{90.7} \\
        
        \bottomrule
      \end{tabular}
  
  \label{3dpw test split}
  \vspace{-10pt}
\end{table}

\noindent
\textbf{3DPW-PC}~\cite{3dpw,romp} is a person-person occlusion subset of 3DPW, comprising 2218 individuals. Images within this dataset contain annotations for multiple persons, potentially distracting the feature extractor. 
This necessitates a robust backbone capable of effectively interpreting the spatial relationships among these individuals. 
As shown in Table~\ref{tab: main experiments}, DPMesh exhibits superior performance compared to previous methods with 82.2 MPJPE and 56.6 PA-MPJPE.

\noindent
\textbf{3DPW-Crowd}~\cite{3dpw,crowdnet} is a person crowded subset of 3DPW
and contains 1923 persons. DPMesh exceeds state-of-the-art methods across all metrics. As demonstrated in Table~\ref{tab: main experiments}, we reach the best result on 79.9 MPJPE and 51.1 PA-MPJPE compared with previous methods.

\noindent
\textbf{CMU-Panoptic}~\cite{cmu} dataset is a multi-person indoor dataset collected with multi-view cameras. In order to ensure a fair comparison, we choose 4 scenes for evaluation, following~\cite{jiang2020coherent,crowdnet}. Results are shown in Table~\ref{cmu panoptic}, and we outshine other competitors on all video clips.

\subsection{Comparisons on Standard Benchmark}
\label{subsec: standard}

\noindent
\textbf{3DPW}~\cite{3dpw} is a widely-used benchmark for human mesh recovery, featuring 60 videos and 3D annotations of 35,515 individuals in its test split. For a fair comparison with other methods, we do not fine-tune our model on the 3DPW train split. As presented in Table~\ref{3dpw test split}, our method achieves state-of-the-art performance on the test split. We also show the qualitative results in Figure~\ref{fig:performance1}, which demonstrate that our DPMesh is robust in complex, wild scenes. 

\subsection{Analysis}
\label{subsec: ablation}

\begin{figure}[t]
  \centering
  \includegraphics[width=0.99\linewidth]{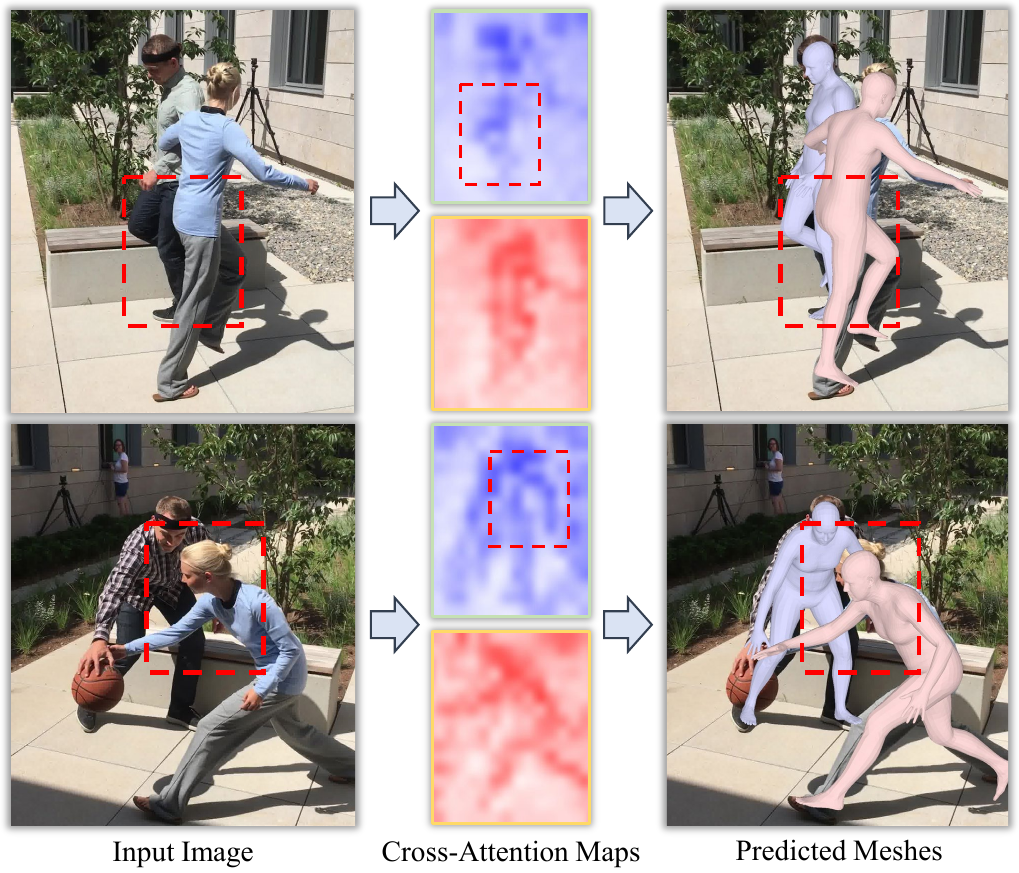}
    \caption{\textbf{Visualization of the cross-attention map.} We demonstrate that DPMesh successfully extracts occlusion-aware knowledge. In the given examples, the male subject is partially occluded by the female subject. The results demonstrate that our DPMesh can precisely capture the spatial relationship and distinguish both targets in cross-attention maps, even when one target is occluded by the other. This enables proficient recovery of occluded meshes.}
  \label{fig:ablation}
  \vspace{-15pt}
\end{figure}

\noindent
\textbf{Effective Backbone with Diffusion Prior.} We engage in a comparative study between our diffusion-based backbone with convolution-based backbones such as ResNet50~\cite{resnet} and HRNet-W48~\cite{hrnet}, as well as transformer-based backbones like ViT-L-16~\cite{vit} and Swin-V2-L~\cite{swinv2}. Note that our selection includes both supervised pre-trained models (\textit{e.g.}, ResNet50) and self-supervised pre-trained models (\textit{e.g}., Swin-V2-L). To implement our comparison, we concatenate pre-detected heatmaps with early-stage image features and fine-tune each backbone for 30 epochs. As revealed in Table~\ref{tab:ablation}, our diffusion-based backbone exhibits superior performance compared to other competitors, proving its exceptional perception capability for occluded human mesh recovery. 
Furthermore, as illustrated in Figure~\ref{fig:ablation}, our diffusion-based backbone accurately captures the occlusion-aware information in the cross-attention maps, which provides explicit guidance for the subsequent regressor by recognizing the target from various occlusions.

\begin{table}[t]
\centering
\caption{ \textbf{Quantitative comparison on CMU-Panoptic~\cite{cmu}}. We measure the MPJPE across different subsets and calculate the overall mean result. DPMesh obtains precise estimation for multi-person scenarios, showing its adaptability to various occlusions.}
\scriptsize
  \vspace{-10pt}
\adjustbox{width=\linewidth}{
      \begin{tabular}{lccccc}
      \toprule
        Method    & Haggl.  & Mafia & Ultim. & Pizza & Mean\\
        \midrule
        {Zanfir \textit{et al.}}~\cite{zanfir2018monocular}  &140.0 & 165.9 & 150.7 & 156.0 & 153.4\\
        {Zanfir \textit{et al.}}~\cite{zanfir2018deep}  &141.4 & 152.3 & 145.0 & 162.5 & 150.3\\
        {Jiang \textit{et al.}}~\cite{jiang2020coherent} & 129.6 & 133.5 & 153.0 & 156.7 & 143.2 \\
        {ROMP}~\cite{romp} & 111.8 & 129.0 & 148.5 & 149.1 & 134.6 \\
        {REMIPS}~\cite{fieraru2021remips} & 121.6 & 137.1 & 146.4 & 148.0 & 138.3\\
        {3DCrowdNet}~\cite{crowdnet} & 109.6 & 135.9 & 129.8 & 135.6 & 127.6\\
        {JOTR}~\cite{li2023jotr} & 99.9 & 113.5 & 115.7 & 123.6 & 114.7\\
        \midrule
        \rowcolor{gray!30}DPMesh (Ours) & \textbf{97.2} & \textbf{109.8} & \textbf{114.3} & \textbf{120.5} & \textbf{110.4}\\
        
        \bottomrule
    \end{tabular}
  }
  
  \label{cmu panoptic}
  \vspace{-5pt}
\end{table}

\begin{table}[t]
	\centering
 \caption{ \textbf{Ablation studies.} We conduct ablations on 3DPW~\cite{3dpw} and 3DPW-OC~\cite{3dpw,3doh} to validate the effectiveness of the diffusion-based backbone, the designed conditions and noisy keypoint reasoning. Notably, unlike other backbones that are pre-trained on perception tasks, we find that DPMesh yields commendable results utilizing a generative pre-trained denoising U-Net. Furthermore, our designed condition injection and noisy keypoint reasoning also enhance overall accuracy.
    }
  \vspace{-10pt}
\large
 \adjustbox{width=\linewidth}
	{\begin{tabular}{lcccc} 
	\toprule 
	\multirow{2}{*}{Settings}& \multicolumn{2}{c}{3DPW} &\multicolumn{2}{c}{3DPW-OC}\\
 \cmidrule(lr){2-3}\cmidrule(lr){4-5}
	&MPJPE$\downarrow$ &PA-MPJPE$\downarrow$  &MPJPE$\downarrow$ &PA-MPJPE$\downarrow$ \\
    \midrule
    \multicolumn{4}{l}{\textit{type of conditioning inputs}} \\
        ${\bm z}_0$&87.6 & 54.9 &89.6 &60.7\\
        ${\bm z}_0+{\bm c}_{\rm j}$&75.1& 49.8&73.7 &50.6\\
        ${\bm z}_0+{\bm c}_{\rm j}^{\rm add}$&100.3& 62.0 &109.0 &73.4\\
       \rowcolor{gray!30}${\bm z}_0+{\bm c}_{\rm j}+{\bm c}_{\rm t}$&\textbf{73.6}& \textbf{47.4}& \textbf{70.9} & \textbf{48.0}\\
    
    \midrule
    \multicolumn{4}{l}{\textit{noisy key-point reasoning}} \\
    {ResNet50~\cite{resnet}~~~~~ w/o NKR}&80.2 & 52.4 &78.9 &52.8 \\
    {HRNet-W48~\cite{hrnet} \! w/o NKR}&78.7 & 50.9 &76.9 &51.8 \\
    {ViT-L-16~\cite{vit} ~~~~~~~ w/o NKR}&76.9 & 49.3 &75.6 &52.0 \\
    {Swin-V2-L~\cite{swinv2}  ~ \! \! w/o NKR}&77.0 & 48.8 &76.1 &52.2 \\
    
    \rowcolor{gray!30}{DPMesh ~~~~~~~~~~~~~~~\!\! w/o NKR}&\textbf{74.9} &\textbf{48.2} &\textbf{73.9}&\textbf{49.9} \\
    \midrule
    \multicolumn{4}{l}{\textit{type of backbones}} \\
    
    {ResNet50~\cite{resnet}}&79.4 & 51.8 &76.1 &50.9 \\
    {HRNet-W48~\cite{hrnet}}&77.2 & 50.8 &75.6 &50.1 \\
    {ViT-L-16~\cite{vit}}&75.2 & 48.5 &73.1 &49.6 \\
    {Swin-V2-L~\cite{swinv2}}&77.3 & 48.5 &73.5 &49.8 \\
    \rowcolor{gray!30}DPMesh (Ours)&\textbf{73.6}& \textbf{47.4}& \textbf{70.9} & \textbf{48.0} \\
	\bottomrule 
	\end{tabular}
 }
  \vspace{-10pt}
    
    \label{tab:ablation}
\end{table}

\noindent
\textbf{Designs of Conditions.} We investigate the effects of different designs for conditioning inputs in our model. The results are presented at the top of Table~\ref{tab:ablation}. We find that both the spatial heatmap condition ${\bm c}_{\rm j}$ and the joint coordinates prompt ${\bm c}_{\rm t}$ can improve performance. This demonstrates the effectiveness of these conditions in helping our model focus on critical areas. Moreover, we experiment with another way to incorporate ${\bm c}_{\rm j}$ by element-wise adding it to the image ${\bm z}_0$. However, this approach (${\bm z}_0+{\bm c}_{\rm j}^{\rm add}$) yields terrible results that are even worse than the baseline without ${\bm c}_{\rm j}$ condition. We assume that simply adding a highly abstracted latent code with the heatmap is meaningless in the latent space and introduces a misleading hint for the model to learn. Consequently, we concatenate ${\bm z}_0$ and heatmap on the channel dimension to reduce mutual interference. 

\noindent
\textbf{Noisy Key-point Reasoning.} We further investigate the effectiveness of our designed NKR approach. As shown in the middle part of Table~\ref{tab:ablation}, we disable NKR on various backbones and evaluate their performance. The results indicate that all backbones without NKR perform worse in occlusion scenes. For the diffusion backbone in DPMesh, NKR approach provides a slight improvement, which confirms its capability to reduce the disturbance of noisy 2D observation. This demonstrates that NKR results in a more robust framework, enabling DPMesh to handle challenging occlusion scenarios and produce more accurate reconstructions.
\section{Conclusion}
In this paper, we introduce DPMesh, a simple yet effective framework for occluded human mesh recovery, which fully exploits the rich knowledge about object structure and spatial interaction within the pre-trained diffusion model. 
We successfully tame the diffusion model with the designed condition injection to perform accurate occluded mesh recovery in a single step. 
Furthermore, we leverage a noisy key-point reasoning approach to enhance the robustness of our model. Extensive experiments demonstrate our framework can achieve accurate estimation even in severe occlusion and crowded environments. We hope our work will provide a new perspective for occluded human mesh recovery and inspire more research in employing diffusion models for perception tasks.

\noindent
\textbf{Acknowledgement.} This work was supported in part by the National Natural Science Foundation of China under Grant 62125603, Grant 62321005, and Grant 62336004, and in part by CCF-Tencent Rhino-Bird Open Research Fund.
{
    \small
    \bibliographystyle{ieeenat_fullname}
    \bibliography{main}
}

\clearpage
\appendix
\section*{Appendix}

\renewcommand\thesection{\Alph{section}}
\setcounter{figure}{0}
\renewcommand\thefigure{\Alph{figure}}
\setcounter{table}{0}
\renewcommand\thetable{\Alph{table}}

In this appendix, we provide additional detailed implementations, qualitative comparisons and ablation studies in Section~\ref{sec: detailed}, Section~\ref{sec: more_qual} and Section~\ref{sec: add_abla}. 

\section{Detailed Implementations}
\label{sec: detailed}
We utilize Stable-Diffusion~\cite{rombach2022high} V1-5 with ControlNet pre-trained specifically on skeleton image conditions so that we can further unleash its diffusion priors. The denoising U-Net has 25 blocks in total, with a middle block and 12 input and output blocks each. The ControlNet takes 12 input blocks as a parallel branch, connected with zero convolution layers as the output layer. The latent code $\bm{z}_0$ is in 32 $\times$ 32 and is downsampled to 16 $\times$ 16 and 8 $\times$ 8 resolutions. During the upsampling process in output blocks, we extract cross-attention maps in the shape of 8 $\times$ 8 and 16 $\times$ 16 and concatenate them with corresponding feature maps in channel dimension. Finally, we use linear layers to merge the pyramid multi-layer feature maps into 2048 $\times$ 8 $\times$ 8 for subsequent regression processing. For the teacher model, we employ the same framework as the student. 

As for the ablation study of various backbones, we follow~\cite{crowdnet,li2023jotr} to initially extract an early-stage image feature in 64 $\times$ 64. Note that for convolution-based backbones (\textit{e.g.}, ResNet50~\cite{resnet} and HRNet-W48~\cite{hrnet}), we use a convolution kernel with 2 strides and a max-pooling layer to downsample the image, while for transformer-based backbones (\textit{e.g.}, ViT-L~\cite{vit} and Swin-V2-L~\cite{swinv2}) we apply patch embedding layers. We set the patch size to 4 for Swin-V2-L and 16 for ViT-L, considering the latter's substantial computational cost. We further quantify the amount of model parameters as illustrated in Table~\ref{tab:comp_param}. With the well-designed implementation of LoRA~\cite{hu2022lora}, we significantly reduce the trainable parameters while effectively maintaining the diffusion prior in the pre-trained models. In comparison with ViT-L, our backbone achieves superior results with reduced computational expenses.

We apply a pose parameter decoder pre-trained on a huge SMPL dataset AMASS~\cite{AMASS:2019}. The framework of our VQVAE is built with linear layers and we take the pose parameters in rotation matrix representation as input in the shape of 24 $\times$ 9. The codebook class number is 2048 and the token dim is 256. During the pre-train stage, we supervise the results with the reconstruction loss following~\cite{SMPL-X:2019}. We also utilize Exponential Weighted Average on codebook optimization inspired by~\cite{Geng23PCT}. As for the inference stage, the regressor will provide 48 tokens to the decoder and finally retrieve the SMPL pose parameters ${\rm \Theta} \in \mathbb{R}^{24 \times 3}$. 

For traninig loss, we set $\lambda_{\rm 2D}$, $\lambda_{\rm 3D}$, $\lambda_{\rm SMPL}$ to 5.0, 2.0 and 1.0 respectively. $\lambda_{\rm NKR}$ is set to 0.1. We speed up training by using distributed training with Pytorch~\cite{paszke2019pytorch} using 8 Nvidia GeForce RTX 4090 GPUs. 

\begin{table}[t]
	\centering
 \caption{ \textbf{Comparison on model parameters.} We quantify the amount of total parameters and trainable parameters across different backbone configurations in the entire pipeline. Given the implementation of LoRA~\cite{hu2022lora}, our DPMesh achieves remarkable results with a minimal fraction of weights finetuned, demonstrating its efficiency and effectiveness.
    }
\large
 \adjustbox{width=\linewidth}
	{\begin{tabular}{lccc} 
	\toprule 
   {Backbone}& {Total Params.}
	&{Trainable Params.} &{MPJPE$\downarrow$} \\
    \midrule
        ResNet50~\cite{resnet} & 39.7M & 39.0M & 76.1\\
        HRNet-W48~\cite{hrnet} & 77.8M & 77.1M& 75.6 \\
        ViT-L~\cite{vit} & 1257.2M & 1256.4M & 73.1 \\
        Swin-V2-L~\cite{swinv2} & 211.2M & 210.3M & 73.5 \\
        \rowcolor{gray!30} DPMesh (Ours) & 1426.3M &408.9M & \textbf{70.9}\\
	\bottomrule 
	\end{tabular}
 }
    \label{tab:comp_param}
\end{table}

\begin{figure*}[t]
  \centering

    \includegraphics[width=0.99\linewidth]{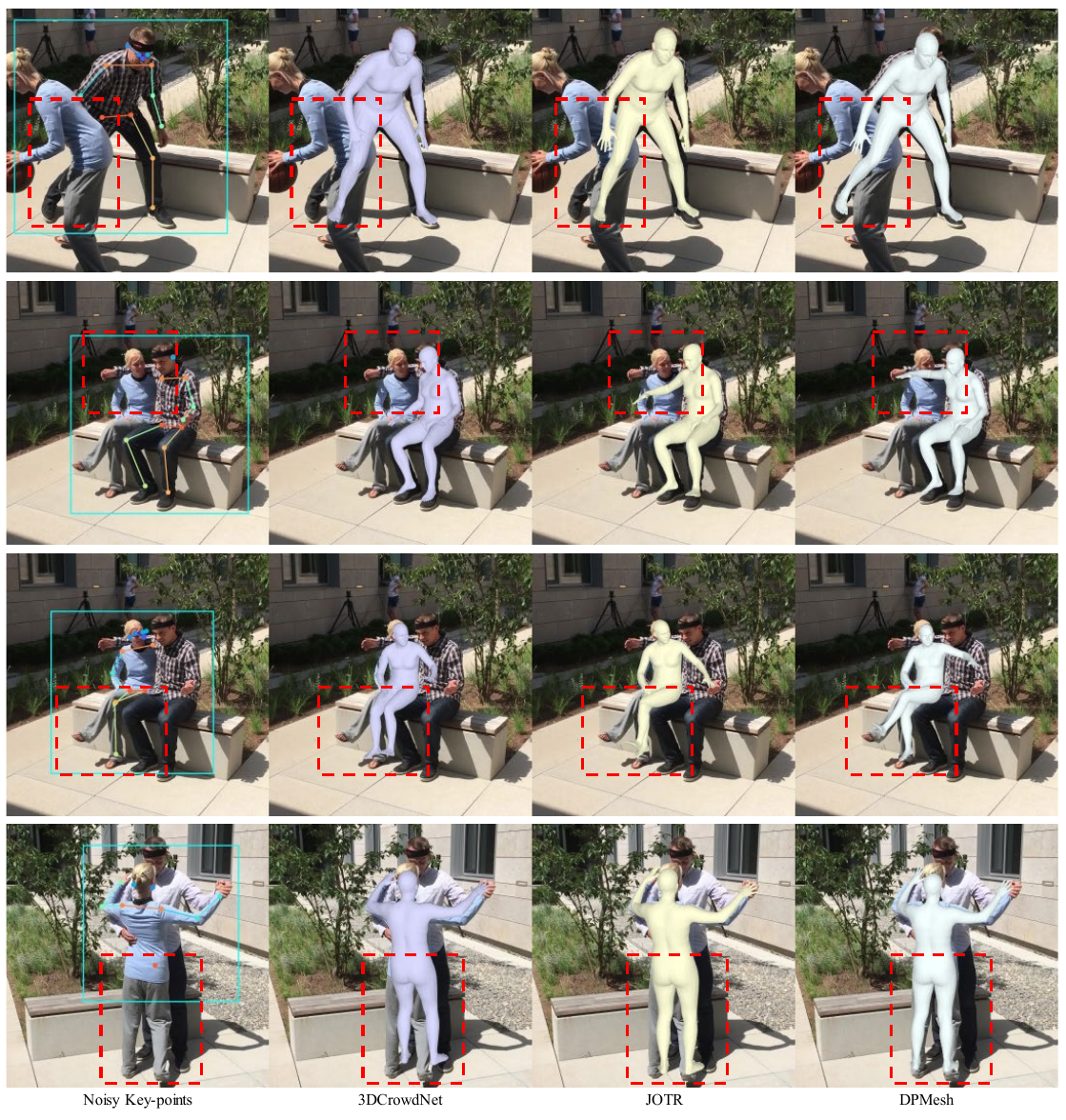}
    \vspace{-10pt}
    \caption{\textbf{Illustration of mesh recovery results under noisy 2D key-points.} In occlusion and crowded scenes, there are 2D key-points missed or inaccurately predicted. Compared with previous methods, our DPMesh with a well-designed Noisy Key-point Reasoning (NKR) approach successfully recovers the correct human mesh.}
  \label{fig:inacc2d}
  
\end{figure*}
\begin{figure*}[t]
  \centering
    \includegraphics[width=0.99\linewidth]{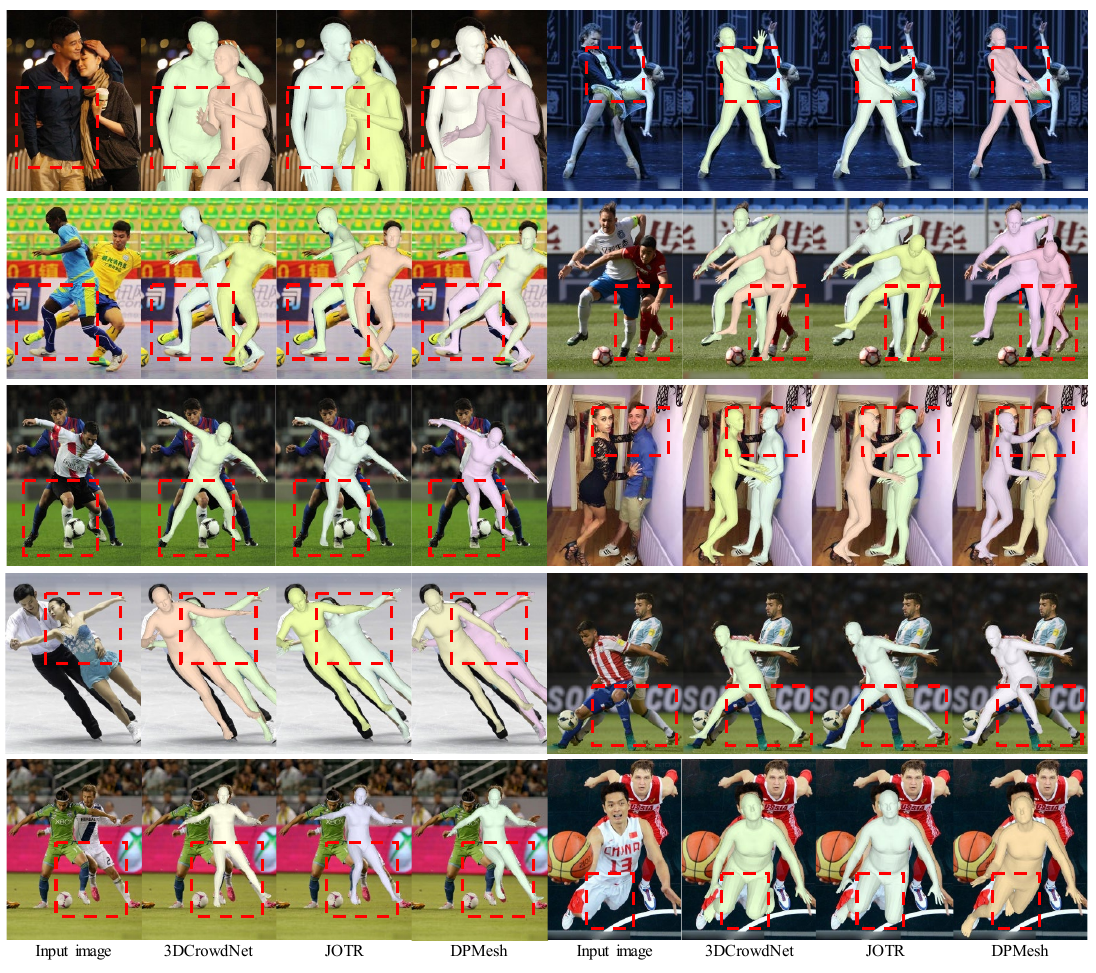}
    \vspace{-10pt}
    \caption{\textbf{Qualitative comparison on OCHuman dataset~\cite{zhang2019pose2seg}.} In more challenging occlusion and crowded scenarios, our DPMesh precisely predicts the human mesh, effectively disregarding interference from adjacent individuals.}
  \label{fig:ochuman}
  
\end{figure*}
\begin{figure*}[t]
  \centering

    \includegraphics[width=0.99\linewidth]{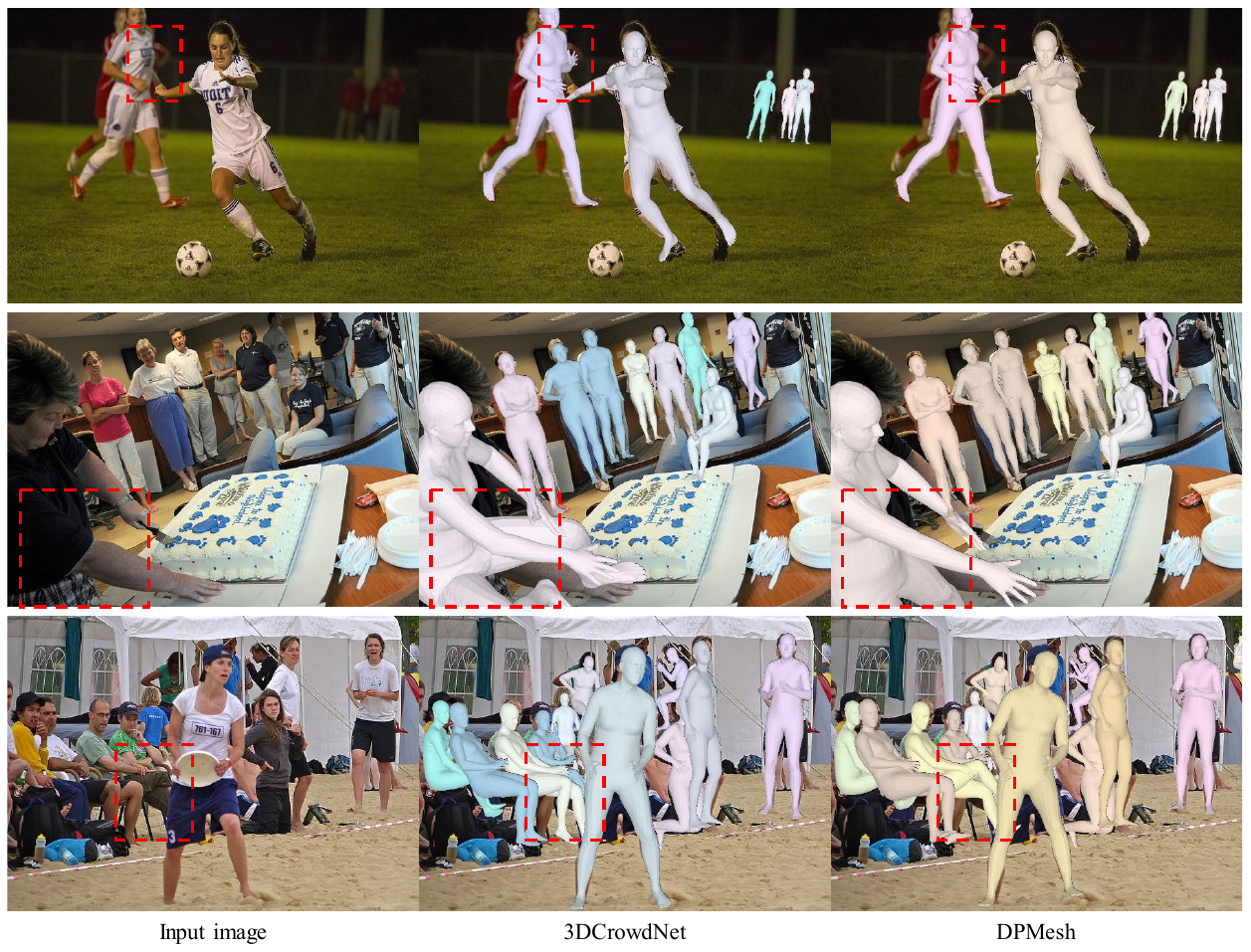}
    \vspace{-10pt}
    \caption{\textbf{Qualitative comparison on CrowdPose dataset~\cite{li2019crowdpose}.} We compare DPMesh with 3DCrowdNet which excels in managing crowded environments. Our DPMesh yields superior outcomes in complex situations such as person-person ambiguity, occlusion with camera distortion and intricate poses within crowds.}
  \label{fig:crowdpose}
  
\end{figure*}

\section{More Qualitative Results}
\label{sec: more_qual}

\noindent
\textbf{Robustness to noisy 2D key-points.} We illustrate the prediction of DPMesh under noisy key-points compared with previous methods in Figure~\ref{fig:inacc2d}. We draw the bounding box according to the region encompassing visible key-points. Given the complexities associated with individual interactions and the absence of precise key-point hints, traditional approaches may yield false predictions. However, by incorporating a robust diffusion backbone and the Noisy Key-point Reasoning (NKR) approach, our DPMesh algorithm achieves a marked improvement in accuracy.

\noindent
\textbf{Comparison on OCHuman~\cite{zhang2019pose2seg}.} We present additional qualitative assessments on OCHuman, an in-the-wild dataset with substantial occlusion consisting of 8,110 meticulously annotated human instances across 4,731 images. Initially, we employ AlphaPose~\cite{alphapose} as an off-the-shelf detector to obtain coarse 2D key-points. Subsequently, we estimate the mesh results with previous methods and our DPMesh. As shown in Figure~\ref{fig:ochuman}, our DPMesh demonstrates exceptional performance in tackling challenging occlusions and complex human poses.

\noindent
\textbf{Comparison on CrowdPose~\cite{li2019crowdpose}.}  The CrowdPose dataset comprises 8,000 images characterized by dense occlusions and complex crowd scenarios. We compare our DPMesh with 3DCrowdNet~\cite{crowdnet}, which is tailored for in-the-wild crowded scenes and addresses 3D human mesh recovery issues with a joint-based regressor. As shown in Figure~\ref{fig:crowdpose}, our DPMesh proficiently estimates the shape and pose of all individuals within the view, effectively handling ambiguous person interactions and body truncations.

\section{Additional Ablation Studies}
\label{sec: add_abla}

\begin{table}[t]
	\centering
 \caption{ \textbf{Ablation study of spatial condition type.} We study the impact of different spatial guidance to denoising U-Net.
    }
\vspace{-10pt}
\large
 \adjustbox{width=\linewidth}
	{\begin{tabular}{lcccc} 
	\toprule 
	\multirow{2}{*}{Conditions}& \multicolumn{2}{c}{3DPW} &\multicolumn{2}{c}{3DPW-OC}\\
 \cmidrule(lr){2-3}\cmidrule(lr){4-5}
	&MPJPE$\downarrow$ &PA-MPJPE$\downarrow$  &MPJPE$\downarrow$ &PA-MPJPE$\downarrow$ \\
    \midrule
        skeleton map&74.9 & 48.4 &72.1 &49.7\\
        heatmap &73.6& 47.4&70.9 &48.6\\
\bottomrule 
\end{tabular}
 }
    \label{tab:abla_skele}
\end{table}
\begin{table}[t]
	\centering
 \caption{ \textbf{Ablation study of different SMPL regressors.} We deploy two distinct types of SMPL regressors  (\textit{e.g.}, recurrent linear layers (RLL)~\cite{kolotouros2019learning} and cascade transformer decoder (CTD)~\cite{li2023jotr}) on both ResNet50~\cite{resnet} and our diffusion-based backbone to investigate their respective performance contributions. 
    }
\large
 \adjustbox{width=\linewidth}
	{\begin{tabular}{lcccc} 
	\toprule 
	\multirow{2}{*}{Settings}& \multicolumn{2}{c}{3DPW} &\multicolumn{2}{c}{3DPW-OC}\\
 \cmidrule(lr){2-3}\cmidrule(lr){4-5}
	&MPJPE$\downarrow$ &PA-MPJPE$\downarrow$  &MPJPE$\downarrow$ &PA-MPJPE$\downarrow$ \\
    \midrule
        ResNet+RLL&81.6 & 53.9 &82.2 &54.1\\
        ResNet+CTD&80.2& 52.4&78.9 &52.8\\
        Diffusion+RLL&75.4& 49.0 &72.5 &50.0\\
        \rowcolor{gray!30}Diffusion+CTD&\textbf{73.6}& \textbf{47.4}& \textbf{70.9} & \textbf{48.0}\\
	\bottomrule 
	\end{tabular}
 }

    \label{tab:abla_spin}
\end{table}
\begin{table}[t]
	\centering
 \caption{ \textbf{Ablation study of LoRA~\cite{hu2022lora} implementation.} We adjust the LoRA rank hyperparameter and unlock more frozen layers in pre-trained diffusion model (\textit{e.g.}, \Checkmark: Weights are unlocked with learnable LoRA matrices. \XSolidBrush: Weights remain consistent with the pre-trained model.). Results are evaluated on 3DPW-OC~\cite{3dpw,3doh}.
    }
\vspace{-10pt}
\scriptsize
 \adjustbox{width=\linewidth}
	{\begin{tabular}{cccc} 
	\toprule 
    {LoRA rank}& {ResBlock}
	&MPJPE$\downarrow$ &PA-MPJPE$\downarrow$ \\
    \midrule
        8& \XSolidBrush &73.2& 50.1\\
        64 & \Checkmark &75.9& 51.5 \\
        \rowcolor{gray!30}64 & \XSolidBrush &\textbf{70.9}& \textbf{48.0}\\
	\bottomrule 
	\end{tabular}
 }
    \label{tab:abla_lora}
\end{table}

\noindent
\textbf{Type of 2D spatial conditions.} In conventional controllable generative models such as ControlNet~\cite{zhang2023adding}, the input human pose guidance is typically provided in the form of a RGB skeleton image detected from Openpose~\cite{cao2017realtime}. Adhering to the approach of ControlNet, we first extract image features from the skeleton image using convolutional layers and then concatenate the spatial feature with ${\bm z}_0$ as a spatial condition. As summarized from Table~\ref{tab:abla_skele}, we assume that the heatmap guidance carries more information such as the joint correspondence to different heatmap channels, than a single skeleton image. Furthermore, noisy key-points may lead to incorrect skeleton connections, which can negatively impact performance. Therefore, our DPMesh utilizes heatmap guidance to effectively introduce spatial conditions.

\begin{table}[t]
 \caption{\textbf{Ablation study on occluded parts}. We conduct an ablation study on our new metrics with cross-attention module (\textbf{CAM}) and Noisy Key-point Reasoning (\textbf{NKR}).}
\vspace{-10pt}
 \adjustbox{width=\linewidth}
	{
 \begin{tabular}{lcccc} 
	\toprule 
	Settings &MPJPE$\downarrow$ &PA-MPJPE$\downarrow$ &OCC-MPJPE$\downarrow$ &OCC-PA-MPJPE$\downarrow$ \\
     \midrule
        JOTR [27] & 75.7 & 52.2 & 89.2 & 63.5  \\
    \midrule
        w/o CAM & 73.7 & 50.6  & 89.4 & 61.0  \\
        w/o NKR & 73.9 & 49.9 & 91.1 & 61.6  \\
    \midrule
        \rowcolor{gray!30}{DPMesh} & \textbf{70.9} &\textbf{48.0} & \textbf{86.2} & \textbf{57.6}  \\
\bottomrule 
\end{tabular}
 }
    \label{tab:rebb_abla_occ}
\end{table}

\noindent
\textbf{Implementation of different mesh regressors.} In order to assess the efficacy of diffusion-based backbone, we apply recurrent linear layers (RLL) derived from~\cite{kolotouros2019learning} and cascade transformer decoder learned from~\cite{li2023jotr} as the SMPL regressor. Results are presented in Table~\ref{tab:abla_spin}. Without bells and whistles, even with vanilla recurrent linear layers, our diffusion-based feature extractor outperforms the performance of JOTR~\cite{li2023jotr}, which carefully designs a contrastive learning loss for its cascade transformer decoder. These findings indicate that our outstanding performance is not severely dependent on the specific regressor employed, highlighting the versatility and effectiveness of the diffusion-based backbone.

\noindent
\textbf{Influence of LoRA~\cite{hu2022lora}.} In order to preserve the diffusion prior within the pre-trained model and minimize computational expenses, we utilize LoRA to finetune only a few parameters in U-Net. We study different LoRA ranks and unlock more blocks to optimize. As shown in Table~\ref{tab:abla_lora}, lower LoRA rank fails to thoroughly unleash the potential of diffusion for visual perception tasks and further unfreezing the ResBlocks may compromise the diffusion prior learned from extensive data. Therefore, employing LoRA matrices to simply unlock cross-attention blocks is appropriate for this specific task, striking a balance between performance and computational efficiency.

\begin{table}[t]
 \caption{\textbf{Comparison with more methods.} We finetune DPMesh with 3DPW training set and compare it with more SOTA methods. DPMesh achieves a competitive result on 3DPW and significantly outperforms them on occlusion benchmark.}
\vspace{-10pt}
 \adjustbox{width=1\linewidth}
	{
 \begin{tabular}{lcccc} 
	\toprule 
	\multirow{2}{*}{Method}& \multicolumn{2}{c}{3DPW} &\multicolumn{2}{c}{3DPW-OC}\\
 \cmidrule(lr){2-3}\cmidrule(lr){4-5}
    &MPJPE$\downarrow$ &PA-MPJPE$\downarrow$  &MPJPE$\downarrow$ &PA-MPJPE$\downarrow$ \\
    \midrule
        HyBrIK~\cite{li2021hybrik} & 71.6 & {41.8} & 90.8 & 58.8\\
        NIKI~\cite{li2023niki}  & 71.3 & \textbf{40.6} & 85.5 & 53.5\\
    \midrule
        \rowcolor{gray!30}{DPMesh-ft}  &\textbf{68.4} & 42.8
        & \textbf{70.9}  & \textbf{48.0}\\
\bottomrule 
\end{tabular}
 }
    \label{tab:supp_comp2}
\end{table}
\begin{table}[t]
 \caption{\textbf{Comparison with HMDiff.} Compared with step-by-step diffusion framework, DPMesh exhibits superior performance, particularly on the 3DPW-PC benchmark.}
\vspace{-10pt}
 \adjustbox{width=\linewidth}
	{\begin{tabular}{lcccc} 
	\toprule 
	\multirow{2}{*}{Method}& \multicolumn{2}{c}{3DPW} &\multicolumn{2}{c}{3DPW-PC}\\
 \cmidrule(lr){2-3}\cmidrule(lr){4-5}
	&MPJPE$\downarrow$ &PA-MPJPE$\downarrow$  &MPJPE$\downarrow$ &PA-MPJPE$\downarrow$ \\
    \midrule
        HMDiff~\cite{hmdiff2023distribution}&72.7& 44.5&114.2 &73.5\\
        \rowcolor{gray!30}{DPMesh (ours)} & \textbf{68.4} & \textbf{42.8} & \textbf{82.2} &\textbf{56.6}\\
\bottomrule 
\end{tabular}
 }
    \label{tab:rebb_hmdiff}
\end{table}

\noindent
\textbf{Evaluation on occluded joints.} Our NKR focuses on dealing with noisy guidance from erroneous key-points. Given that these noisy key points only take a small fraction of the total body, they may not have a significant impact on the overall result. To further assess the NKR’s impact, we introduce \textbf{OCC-MPJPE$\downarrow$} and \textbf{OCC-PA-MPJPE$\downarrow$} to evaluate errors on occluded joints. As shown in Table~\ref{tab:rebb_abla_occ}, the cross-attention module and the Noisy Key-point module both are effective on occluded key-points input.

\noindent
\textbf{More comparisons.} We compare DPMesh with more methods. For a fair comparison, when testing on the 3DPW test split, we fine-tune our model with the 3DPW~\cite{3dpw} training set. As illustrated in Table~\ref{tab:supp_comp2}, DPMesh achieves competitive performance on the 3DPW test split benchmark and significantly outperforms the occlusion benchmark. Furthermore, we compare our one-step DPMesh with the step-by-step denoising framework HMDiff~\cite{hmdiff2023distribution}. HMDiff is an optimization method that takes over 200 steps to recover human mesh. As shown in Table~\ref{tab:rebb_hmdiff}, DPMesh exhibits much better results on the 3DPW-PC benchmark while also outperforming HMDiff on the 3DPW test split.

\end{document}